\pdfoutput=1

\documentclass[11pt]{article}

\usepackage[]{acl}

\usepackage{times}
\usepackage{latexsym}

\usepackage[T1]{fontenc}

\usepackage[utf8]{inputenc}

\usepackage{microtype}

\usepackage{inconsolata}

\usepackage{times}
\usepackage{latexsym}
\usepackage{amsmath}
\usepackage{amssymb}
\usepackage{graphicx}

\DeclareMathOperator*{\argmax}{argmax}

\def\rouge{34}
\def\intent{89}
\def\humaneval{86}

\usepackage{multirow}
\usepackage{url}
\usepackage{algorithm}
\usepackage{algpseudocode}
\usepackage{booktabs}
\usepackage{comment}
\usepackage{bbm}
\usepackage{breqn}

\usepackage[most]{tcolorbox}

\tcbset{
    frame code={},
    center title,
    left=0pt,
    right=0pt,
    top=0pt,
    bottom=0pt,
    colback=gray!70,
    colframe=white,
    width=\dimexpr\textwidth\relax,
    enlarge left by=0mm,
    boxsep=5pt,
    arc=0pt,outer arc=0pt,
    }

\usepackage{breakurl}
\usepackage[breaklinks]{hyperref}

\newcommand\revision[1]{\textcolor{black}{#1}}

\usepackage{pifont}

\title{Enhancing AI Assisted Writing with One-Shot Implicit Negative Feedback}

\author{Benjamin Towle\textsuperscript{1}, Ke Zhou\textsuperscript{1,2} \\
  \textsuperscript{1}University of Nottingham \\
  \textsuperscript{2}Nokia Bell Labs \\
  \texttt{\{benjamin.towle, ke.zhou\}@nottingham.ac.uk} \\\
}

\begin{document}
\maketitle
\begin{abstract}
AI-mediated communication enables users to communicate more quickly and efficiently. Various systems have been proposed such as smart reply and AI-assisted writing. Yet, the heterogeneity of the forms of inputs and architectures often renders it challenging to combine insights from user behaviour in one system to improve performance in another. In this work, we consider the case where the user does not select any of the suggested replies from a smart reply system, and how this can be used as \textit{one-shot implicit negative feedback} to enhance the accuracy of an AI writing model. We introduce \textsc{Nifty}, an approach that uses classifier guidance to controllably integrate implicit user feedback into the text generation process. Empirically, we find up to \rouge\% improvement in \textsc{Rouge-L}, \intent\% improvement in generating the correct intent, and an \humaneval\% win-rate according to human evaluators compared to a vanilla AI writing system on the MultiWOZ and Schema-Guided Dialog datasets. The code is available at \url{https://github.com/BenjaminTowle/NIFTY}.
\end{abstract}

\section{Introduction}
The average worker reportedly spends around 23\% of their time on reading and answering emails \cite{Mark2012APN}. To alleviate this burden, there is a growing demand for AI-mediated communication systems to draft and potentially fully automate replies for users. These facilitate faster communication by providing suggestions at different stages of the conversational pipeline. Various modes of interaction exist, each with differing trade-offs: smart reply systems -- such as in Gmail \cite{Henderson2017EfficientNL} or Outlook \cite{Deb2019DiversifyingRS} -- offer a low-latency solution to dealing with simple requests, using a retrieval-based model to present canned suggested replies to the user which can be clicked instead of requiring manual typing. AI writing models -- such as used by \href{https://www.jasper.ai/}{Jasper} or \href{https://www.grammarly.com/}{Grammarly} -- employ generative architectures to produce more complex replies, but may require additional manual editing and / or prompting from a user to obtain the desired result. The smart reply system may also provide an initial skeleton reply, that an AI writing model can later improve upon  \citep{Chen2019GmailSC}. 

\begin{figure}
    \centering
    \includegraphics[scale=0.1]{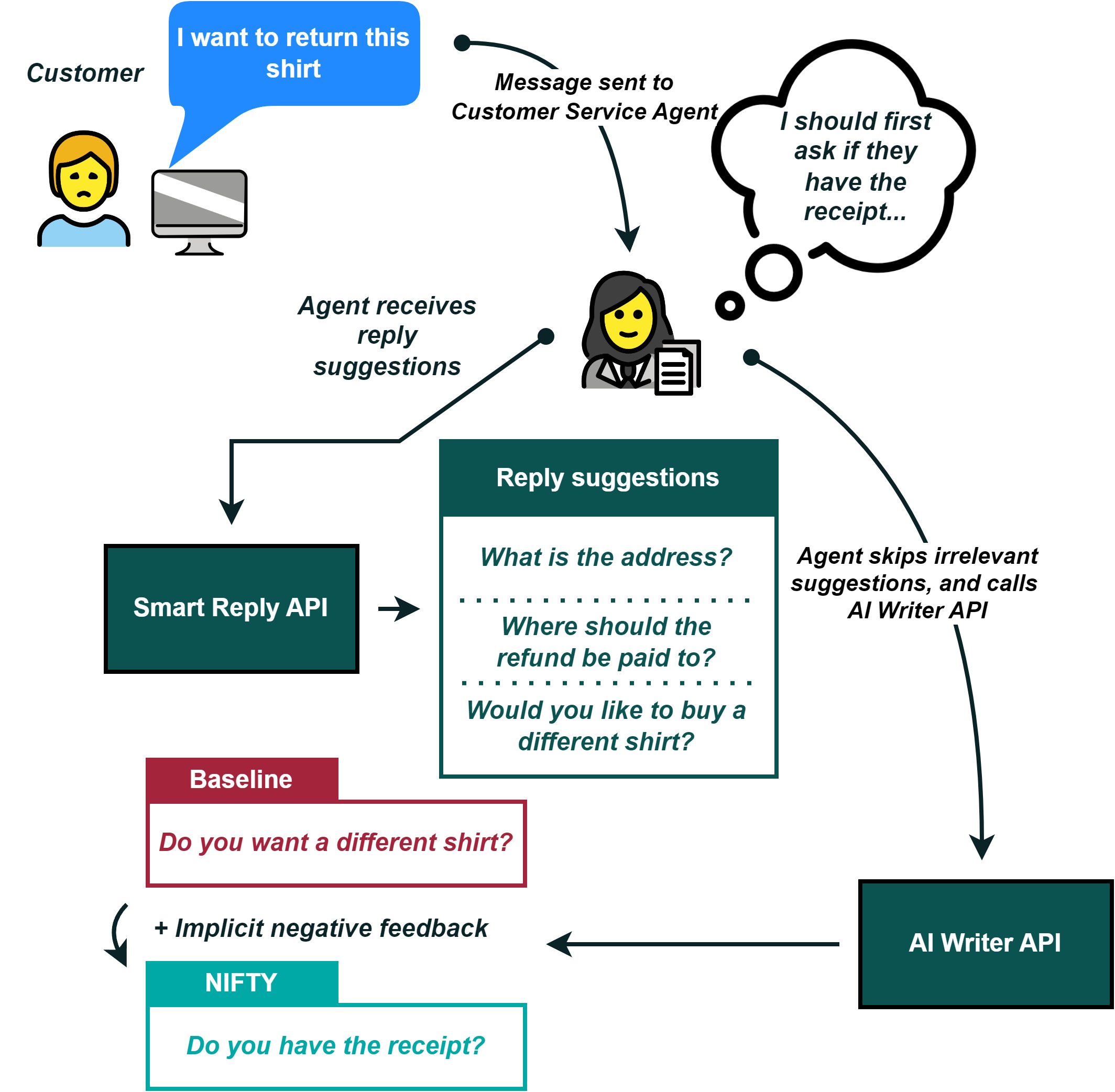}
    \caption{Example of how an agent may utilise either a smart reply system or an AI writing system to speed up communication with a customer. Our approach, \textsc{Nifty}, uses implicit negative feedback from the rejected suggestions to improve the AI writer's prediction.}
    \label{fig:pipeline}
\end{figure}

Yet, the heterogeneity of the forms of interaction and architectures often renders it non-obvious how the information from one system can be leveraged to improve another: e.g., a smart reply system is only useful when the user clicks one of the suggestions. In practice however, there is often too much uncertainty surrounding the user's intent for any of the suggestions to be relevant (Figure \ref{fig:pipeline}) \cite{linkedinSR}. Due to reading the suggestions, this increases the user's cognitive load, with no additional payoff. This contrasts with positive feedback, when selecting a suggestion reduces user typing time. 

To address this problem, we conduct a pilot study into how \textit{one-shot implicit negative feedback} can be used to integrate and improve the performance between two heterogeneous AI-mediated communication systems. In particular, we consider how feedback when a user clicks \textit{none} of a smart reply system's suggestions can improve a downstream AI writing model at run-time. We concentrate on the one-shot setting, which has the advantage of enabling a single shared model for all users, as well as being more challenging due to the limited amount of interaction information per user. Future work may extend this to greater degrees of user personalisation, although this is currently out of scope due to lack of data access.

In this paper, we introduce \underline{\textbf{N}}egative \underline{\textbf{I}}mplicit \underline{\textbf{F}}eedback from Smar\underline{\textbf{t}} Repl\underline{\textbf{y}} ("\textsc{Nifty}"). \textsc{Nifty} employs classifier guidance \cite{yang-klein-2021-fudge} to controllably integrate one-shot implicit negative feedback into the generation process at run-time. In particular, given an unconditional AI writing model, we condition the model on a desired attribute $c$, via an application of Bayes' rule, using a classifier trained to predict $c$. We explore two possible settings for $c$: an intent-based approach that conditions on the most likely next intent not represented in the suggestions and a user action-based approach that conditions directly on the user rejecting the suggestions. Overall our method affords several key advantages: (i) it keeps the smart reply and AI writing systems decoupled, allowing it to be readily integrated into existing systems which can be optimised by separate teams; (ii) it enables additional forms of negative feedback to be introduced in the future, e.g., lingering on a suggestion without clicking it \citep{Zhang2015adaQACAQ}, via a linear combination with a separate classifier; (iii) incorporating implicit negative feedback into the AI writing process reduces the cost to the user of being presented with irrelevant suggested replies.

Empirically, by evaluating on two publicly available datasets, we find up to \rouge\% improvement in \textsc{Rouge-L} and \intent\% improvement in generating the correct intent compared to a vanilla AI writing system, and an \humaneval\% win-rate according to human evaluators. In summary, our key contributions are: (1) We introduce the framework of implicit negative feedback to the smart reply and AI writing tasks; (2) We develop and open-source an approach that uses classifier guidance, considering both intent-based and user action-based attributes for conditioning; (3) We provide both quantitative and qualitative analysis of our model's superior performance using both automated and human evaluation.

\section{Related Work}
Early AI-mediated communication centred around smart reply systems, which provided canned replies to a user message \cite{Kannan2016SmartRA, Henderson2017EfficientNL}. Work then expanded to various forms of heterogeneous interaction including AI writing systems, such as autocompletion \cite{Chen2019GmailSC}, alternative word suggestions \cite{wang2023smartwordsuggestions}, or conditioning on: rough drafts \citep{ito2019diamondsinthe}, abbreviated sentences  \citep{adhikary2021acceleratingtextcommunication}, or user provided intents \citep{sun2021igaanintent}. While these methods all require explicit user effort, our approach requires no interactive effort from the user, by leveraging implicit negative feedback.

Various types of implicit feedback have been explored in previous works, such as user lingering time \citep{Zhang2015adaQACAQ}, skipping content \citep{Pan2023LearningAO, Gong2022PositiveNA}, or conversation length \citep{Irvine2023RewardingCF}. To the best of our knowledge, our work is the first to apply this to the smart reply setting.  

Finally, our work builds upon the growing literature on controllable text generation. Here, early work focused on conditioning generation on particular control `codes' \cite{Keskar2019CTRLAC}. More recently, work focuses on classifier guidance, in which a separate classifier guides token-by-token generation \citep{Krause2020GeDiGD, yang-klein-2021-fudge, Shuster2021AmIM, Arora2022DirectorGF}. Note that our focus is \textit{not} to introduce a novel algorithm for classifier guidance, but rather to demonstrate how it can be combined with implicit negative feedback to solve a problem in AI-mediated communication -- namely -- the lack of easy integration between different modes of interaction. Therefore, as new techniques emerge from this field they may be used to enhance our own method.


\section{Method}
\paragraph{Filtering with User Simulator}
We run our user simulator using the dataset $\mathcal{D} = \{(m,r,\mathbf{s})\}_{i=1}^I$, where $m$ is the incoming message, $r$ is the ground-truth reply, and $\mathbf{s} = \{s_1,...s_K\}$ is the set of reply suggestions obtained from a black-box smart reply system. We are concerned with the scenario in which the user clicks \textit{none} of these suggestions. To represent this concretely, let $\mathbf{z}$ represent the set of all possible intents, $\mathbf{z_s}$ be the set of intents assigned to suggestions in $\mathbf{s}$, and $z$ be the intent of the ground-truth response $r$ -- e.g., \textit{`Yes I can! Table for 1?'} corresponds to the \texttt{booking-inform} intent. We simulate the user by having the user reject all suggestions when $z \not\in \mathbf{z_s}$, i.e. none of the suggestions contain the intent of the ground-truth reply. By filtering according to this criterion, we obtain dataset $\mathcal{D}'$ which is used for downstream evaluation. Note that both our generator and classifier are trained on the full version of our dataset $\mathcal{D}$.

\paragraph{Smart Reply}
The smart reply system uses a vector retrieval model \citep{Karpukhin2020DensePR}, trained to jointly embed messages and replies into a shared latent space. At run-time, it retrieves the top $K$ nearest neighbours as suggested replies. Following convention \citep{Deb2019DiversifyingRS}, we set $K = 3$ for all of our experiments. We choose this straightforward approach in order to demonstrate that even an out-of-the-box smart reply system can provide useful implicit negative feedback, without requiring any bespoke alterations.\footnote{While there are at least two open-source implementations for this \citep{Zhang2021ADA, towle-zhou-2023-end}, we use the latter due to its native support for adding new datasets.} 

\paragraph{Generator}
Following previous approaches \citep{sun2021igaanintent, faltings2023interactivetextgeneration}, the AI writing model is a transformer trained to generate tokens autoregressively. Given a message $m$, the probability of generating reply $r$ can be factorised as:
\begin{equation}
\label{eq:cg}
\mathbf{p}_\Theta(r|m) = \prod_{t=1}^T \mathbf{p}_\Theta(r_t|m,r_{<t})
\end{equation}
This model is trained only on the (message, reply) pairs from $\mathcal{D}$, without requiring access to the smart reply system, or any implicit user feedback, using negative log likelihood.

\paragraph{Classifier Guidance}
allows an unconditional text generation model $\mathbf{p}_\Theta(r_t|m, r_{<t})$ to generate tokens conditioned on an attribute $c$, by performing a classification step over possible next tokens $\mathbf{p}_\Phi(c | m, r_{\leq t})$. These can be combined through Bayesian decomposition \citep{yang-klein-2021-fudge}.
\begin{equation}
\resizebox{.8\hsize}{!}{$\hat{\mathbf{p}}(r_t|m, r_{<t}, c) \propto
    \mathbf{p}_\Theta(r_t|m, r_{<t}) \cdot \mathbf{p}_\Phi(c | m, r_{\leq t})$}
\end{equation}
Note, the classifier predicts whether the attribute \textit{will} be obtained by the time the response is completed, not whether the attribute is currently present. By operating at the token level, rather than over completed generations (i.e. reranking), the model is able to explore a larger search space.

For our purposes, we consider two possible approaches for what attribute to condition the generator on. First, we condition on the desired intent of the response. In particular, we train a classifier to predict the final intent of the reply, given only the reply prefix $r_{\leq t}$ and message $m$. At run-time, we then condition on the most likely intent that is not present in the suggestions:
\begin{equation}
    \argmax_{j=1:J} \mathbf{p}_\Phi(z_j|m,r_{\leq t}) \cdot \mathbbm{1}_{\mathbf{z} \setminus \mathbf{z_s}}(z_j)
\end{equation}
where $\mathbf{z} \setminus \mathbf{z_s}$ is the set of intents not present in the suggestions, and $\mathbbm{1}$ is the indicator function that outputs 1 if $z_j \in \mathbf{z} \setminus \mathbf{z_s}$, 0 otherwise. The main limitation of this approach is that it requires access to a labelled dataset of intents. To remove this limitation, we therefore also consider conditioning the classifier directly on the user action. Specifically, we attempt to predict whether or not the user rejected the suggested replies as a binary classification task.

\section{Experiment}
\paragraph{Datasets} We evaluate our method on two task-oriented datasets covering various domains: MultiWOZ v2.2 \citep{Budzianowski2018MultiWOZA} and Schema-Guided Dialog (SGD) \citep{dstc8}. Both have the advantage of being annotated with intents, and also feature the professional style of conversations that many AI writing applications aim to facilitate. We treat replies containing multiple intents as unique intents in their own right as it is unclear that a user would accept a suggestion that only partially contained the desired intents. As we are concerned with the assisted writing setting, rather than creating a consumer-facing task-oriented chatbot, we evaluate using standard AI writing metrics, rather than success-based task-oriented metrics. See Appendix \ref{sec:appendix_datasets} for dataset statistics.

\paragraph{Metrics}
We employ both automatic and human evaluation. \textsc{Rouge-L}, which has been used previously in AI writing tasks \citep{ito2019diamondsinthe}, measures the longest common subsequence within the prediction, capturing surface-level overlap with the ground-truth. However, there are often many ways of expressing the same meaning that lack term-overlap. Therefore, inspired by previous evaluation work on smart reply systems \citep{Weng2019OCCAS}, we complement this metric with R@1, which measures the proportion of generated responses that contain the correct intent. As the responses are generated on-the-fly, we detect intents using a DistilBERT-based classifier trained to map utterances to the corresponding intent for each dataset (a separate classifier to the one used in \textsc{Nifty}), which we then compare to the ground-truth intent.  We conduct human evaluation using Amazon Mechanical Turk. As the purpose of the system is to reduce uncertainty about the user's intended reply, we apply a pairwise setup, in which the evaluator is asked to \textit{select which candidate reply is most similar to the target reply}. For each model and dataset, we randomly select 100 data points from the test set and assign 3 annotators to each data point. We define a `win' as when a system is achieves a majority vote for a given data point. \revision{Overall, the procedure was carried out by 268 unique annotators, across 400 data points. See Appendix \ref{sec:amt} for further details.}

\begin{table}[t]
\centering
\small
\begin{tabular}{llllll}
\toprule
    & & \multicolumn{2}{c}{MultiWOZ} & \multicolumn{2}{c}{SGD} \\
    \cmidrule(lr){3-4} \cmidrule(lr){5-6}
    Base & Method & R-L & R@1 & R-L & R@1 \\
    \midrule
    \multirow{7}{*}{BB} & Baseline & 25.0  & 15.6 & 16.7 & 28.2 \\
    & Unlikelihood & 25.3 & 15.0 & 16.2 & 29.1 \\
    & Rules-based & 25.2 & 17.5 & 16.6 & 25.1  \\
    & Reranker Action  & 24.9  & 15.4 & 16.7 & 28.7 \\
    & Reranker Intent & 25.2 & 17.9 & 16.8 & 29.2 \\
    \cmidrule(lr){2-6}
    & \textsc{Nifty} Action & 25.3  & 18.1 & 20.4 & 43.2  \\
    & \textsc{Nifty} Intent & \textbf{27.2}* & \textbf{26.1}* & \textbf{21.6}* & \textbf{51.2}* \\
    \midrule
    \multirow{7}{*}{T5} & Baseline & 27.7 & 16.4 & 18.0 & 28.1 \\
    & Unlikelihood & 25.3 & 11.1 & 17.3 & 26.1 \\
    & Rules-based & 27.8 & 18.0 & 20.2 & 30.7\\
    & Reranker Action  & 24.8 & 12.8 & 19.7 & 35.9 \\
    & Reranker Intent & 27.6  & 17.1 & 19.6 & 34.8  \\
    \cmidrule(lr){2-6}
    & \textsc{Nifty} Action & 24.8 & 12.7 & 23.8 & 49.6 \\
    & \textsc{Nifty} Intent & \textbf{29.0}* & \textbf{28.5}* & \textbf{24.2}* & \textbf{53.2}* \\
    \bottomrule
\end{tabular}
\caption{Results on MultiWOZ and SGD test sets using \textsc{Rouge-L} and R@1 metrics. \textbf{Bold} indicates best result. * indicates result is statistically significant with $p$-value $< 0.01$ compared to best baseline.}
\label{tab:main}
\end{table}

\paragraph{Baselines} We compare our approach to: the standard \textit{Baseline} AI writing model -- i.e. without any classifier guidance; a \textit{Reranker} approach which reranks the final output beams, without any token-level reranking, by selecting the beam with the highest score for the desired intent / action; an \textit{Unlikelihood} decoding approach that downweights the probability for terms that occur in the rejected intents, encouraging the model to generate one of the non-rejected intents \citep{unlikelihood}; \revision{a \textit{Rules-based} approach in which multiple candidate beams are generated, and are then filtered to remove beams containing rejected intents from the reply suggestions.}

\paragraph{Model Details}
We explore two different transformers for generation: BlenderBot-400M (BB) \citep{Roller2020RecipesFB} and T5-220M (T5) \citep{Raffel2019ExploringTL}. At run-time, we generate responses using beam search ($n=5$). For efficiency, we use the lightweight DistilBERT-66M for classification \citep{Sanh2019DistilBERTAD} and only rescore the top $10$ tokens with it, following \citep{Shuster2021AmIM}. \revision{All models are trained with a batch size of 32, learning rate of 5e-5 with linear decay until convergence using the AdamW optimiser. For the generative models, we train using low-rank adaptors (LoRA) \citep{Hu2021LoRALA}, with an $r$ value of 8, alpha of 32 and dropout of 0.1, see Appendix \ref{sec:appendix_hyperparameters}}.

\begin{table}
\centering
\small
\begin{tabular}{llrrrr}
\toprule
    & & \multicolumn{2}{c}{MultiWOZ} & \multicolumn{2}{c}{SGD} \\
    \cmidrule(lr){3-4} \cmidrule(lr){5-6}
    Model & $\alpha$ & R-L & R@1 & R-L & R@1 \\
    \midrule
    \multirow{3}{*}{T5 \textsc{Nifty} Intent} & 0.5 & 29.4 & 27.8 & 24.5 & 53.1 \\
    & 1.0 & 29.0 & 28.5 & 24.2 & 53.2 \\
    & 2.0 & 27.9 & 28.9 & 24.0 & 53.5 \\ 
    \bottomrule
\end{tabular}
\caption{\revision{Results on MultiWOZ and SGD test sets under different values of $\alpha$, using the T5 \textsc{Nifty} Intent model.}}
\label{tab:alpha}
\end{table}

\begin{table}[t]
    \centering
    \small
    \begin{tabular}{lllrr}
    \toprule
    & & & MultiWOZ & SGD \\
    \cmidrule(lr){4-4} \cmidrule(lr){5-5}
        Base & Winner & Loser & Win rate & Win rate \\
        \midrule
          BB & \textsc{Nifty} Intent & Baseline & 73\%* & 77\%* \\
          T5 & \textsc{Nifty} Intent & Baseline & 86\%* & 78\%*  \\
         \bottomrule
    \end{tabular}
    \caption{Human evaluation using crowdworkers. * indicates result is statistically significant with $p$-value $< 0.01$ using binomial test.}
    \label{tab:human}
\end{table}

\begin{table*}
\centering
\small
\begin{tabular}{l|l}
\toprule
     Message & \textit{Sure, 2 people, checking in on Thursday for 3 nights please.}  \\
     \midrule
     Sugg. \#1 & \texttt{[Booking-NoBook]} \texttt{[general-reqmore]} \\
     & Booking was unsuccessful. Would you like to find another hotel? \\
     Sugg. \#2 & \texttt{[Booking-NoBook]} \\
     & Booking was unsuccessful. \\
     Sugg. \#3 & \texttt{[general-reqmore]} \\
     & Will you be needing a reference number? \\
     \midrule
     Baseline & I'm sorry, I can't make that for you. Would you like me to try another time? \\
     \midrule
     \textsc{Nifty} & Your booking was successful. Your reference number is \#\#\#\#. Can I help you with anything else? \\
     \midrule
     Target & \texttt{[Booking-Book]} \texttt{[general-reqmore]} \\
     & Your booking is successful! Your reference number is \#\#\#\# . Can I help you with anything else? \\
     \bottomrule
\end{tabular}
\caption{Qualitative example from the MultiWOZ test set. Text is post-processed to remove reference numbers as the model does not have access to booking API. Text in square brackets is annotated intents from dataset.}
\label{tab:qualitative}
\end{table*}

\paragraph{Main Results}
Table \ref{tab:main} presents the results for our overall system. We find our approach consistently increases the model's ability to generate the correct intent, with an improvement of up to \intent\% compared to the baseline. This further corresponds to an improvement in \textsc{Rouge-L} of up to \rouge\% compared to the baseline. We find that the Reranker \revision{and Rules-based approaches} fail to improve much upon the Baseline approach, which is consistent with the findings of previous work \citep{Shuster2021AmIM}, as the resulting beams typically do not represent a broad range of intents. The Unlikelihood approach also struggles due to different intents often having significant term overlap. Choice of base model proved important for some methods, although ultimately \textsc{Nifty} Intent remained the strongest approach in both cases. In terms of choice of classifier, we find the intent-based classifier performs significantly better, especially for MultiWOZ. This gap is much narrower in SGD however, which we hypothesise is due to the smaller number of intents making it easier for the Action classifier to implicitly learn them. Future work may investigate techniques such as unsupervised intent discovery, which has already been used in smart reply to replace the reliance of \textsc{Nifty} Intent on labelled data \citep{Kannan2016SmartRA}.

We note also that the classifier used for intent prediction given the message has an R@1 of 47.4\% for MultiWOZ and 83.2\% for SGD. This difference is due to MultiWOZ having significantly more possible intents (see Appendix \ref{sec:appendix_datasets}). Future work may explore techniques to improve accuracy such as increasing the context window to multi-turn, as we expect this to improve the overall task performance.

\paragraph{Varying Degree of Guidance}
In Table \ref{tab:alpha}, we further evaluate how performance for the strongest version of our method, T5 \textsc{Nifty} Intent, varies under different levels of $\alpha$, which determines the weighting of the classifier on the logits, i.e., the rightmost term of Equation \ref{eq:cg}. We find although the approach is broadly flexible to a range of values, higher \textsc{Rouge-L} scores are associated with weaker levels of guidance, while stronger levels result in higher R@1. Overall, we find the middle-ground setting of 1.0 offers the best trade-off, as well as not requiring any additional hyperparameter search.

\paragraph{Human Evaluation}
Table \ref{tab:human} shows our human evaluation results. Across both datasets and models, \textsc{Nifty} Intent is statistically significantly judged better in the pairwise evaluation, with up to an 86\% win rate.

\paragraph{Case Study}
Table \ref{tab:qualitative} presents a qualitative example of model performance. By utilising the fact that the user simulator rejected the unsuccessful booking intent, \textsc{Nifty} correctly surmises that a successful booking intent is instead required, in contrast to the baseline model, which refuses to make the booking. \revision{This example also supports our intuition in designing the user simulator on the assumption that the user would not select a suggestion that only partially overlapped in intent with the ground truth. In this case, although \texttt{[general-reqmore]} is present in both suggestions and target, the lack of the \texttt{[booking-book]} intent appears critical to appropriately responding to the message}.

\section{Conclusion}
In this work we introduce \textsc{Nifty}, an AI writing system that uses classifier guidance to account for implicit negative user feedback from an upstream smart reply system. Empirically, we find up to \rouge\% improvement in relevance and \intent\% improvement in generating the correct intent compared to a vanilla AI writing system. Future work may explore applying these techniques to real-life data from deployed systems and / or may be extended to a broader range of types of feedback beyond click data from smart reply systems.

\section*{Acknowledgements}
We thank the reviewers for their helpful feedback and suggestions. This work is partly supported by the EPSRC DTP Studentship program. The opinions expressed in this paper are the authors', and are not necessarily shared/endorsed by their employers and/or sponsors.

\section*{Limitations}
We identify three main limitations of our work, which we address here. First, our work assumes the main driver of a user clicking a suggestion is whether it matches their intended intent. However, there may be additional factors that influence which suggestions a user clicks, such as formality of the utterance, the user's own preference for using AI generated content, or the user's own writing style. However, deployed systems may circumvent this issue by training directly on user click through data, in place of the user simulator used here. Second, the datasets used are somewhat artificial, being deliberately designed as dialogue benchmarks, rather than organic datasets created in an actual working environment, which are potentially more noisy. Third, the most effective version of our method, \textsc{Nifty} Intent requires a dataset of intent annotations to train the classifier. We regard removing this limitation, such as via unsupervised intent detection, as a possible avenue for future work.

\section*{Ethical Considerations}
Various ethical concerns have been raised around the potential for generative dialog systems to produce inappropriate content, particularly as their fluency increases and their content because less distinguishable from a human's. However, there is additional nuance in the context of AI-assisted writing in that humans have oversight over the content being generated, and may reject it if it is inappropriate. On the one hand, this may be seen as a mitigating factor, as the human may function as an implicit safety classifier. On the other hand, recent research indicates that suggestions may influence user behaviour to a degree; specifically, while smart reply systems have long been known to have a positivity bias \citep{Kannan2016SmartRA}, recent work finds that this can influence the behaviour of the system's users. In particular, \citet{Wenker2022WhoWT} find that users of smart reply systems produced overly more positive sentiment replies than users without access to these systems. Further research is needed on understanding in what other ways assisted writing systems shift the distribution of user replies.

\bibliography{local}

\appendix

\section{Dataset Details}
\label{sec:appendix_datasets}
Table \ref{tab:datasets} presents the detail for the MultiWOZ and SGD datasets. We filter the datasets to include only those examples where the suggestions were rejected in the user simulator, i.e. none of the suggestions had a shared intent with the ground-truth reply. The datasets contain a large number of intents, as we treat replies containing multiple intents as standalone intents.

\begin{table*}[]
    \centering
    \begin{tabular}{lcccc}
        \toprule
        & \multicolumn{2}{c}{MultiWOZ} & \multicolumn{2}{c}{SGD} \\
        \cmidrule(lr){2-3} \cmidrule(lr){4-5}
        & Pre-filtering & Post-filtering & Pre-filtering & Post-filtering \\
        \midrule
       Train size  & 56.8k & 36.9k & 165.0k & 30.2k \\
       Validation size & 7.4k & 4.9k & 24.4k & 5.2k \\
       Test size & 7.4k & 4.9k & 42.3k & 9.8k  \\
       \midrule
       \# Intents & \multicolumn{2}{c}{685} & \multicolumn{2}{c}{21} \\
       \midrule
    \end{tabular}
    \caption{Statistics for the MultiWOZ v2.2 \cite{{Budzianowski2018MultiWOZA}} and SGD \cite{dstc8} datasets, indicating number of samples before and after filtering to include only samples with negative feedback (i.e., rejected smart replies). Number of intents is large as multi-intent replies are treated as unique intents.}
    \label{tab:datasets}
\end{table*}

\section{Training Hyperparameters}
\label{sec:appendix_hyperparameters}
Table \ref{tab:hyperparameters} summarises the training hyperparameters. To reduce training time for the generator, we use LoRA \citep{Hu2021LoRALA}. All classifiers used in the paper use the same hyperparameters. We train the classifier for more epochs than the generator as the classifier requires learning the weights for the individual classes in the classifier head from scratch. All remaining unstated hyperparameters are the default provided by the HuggingFace \texttt{TrainingArguments} class.
\begin{table*}[]
    \centering
    \small
    \begin{tabular}{lcccccc}
    \toprule
    & \multicolumn{3}{c}{MultiWOZ} & \multicolumn{3}{c}{SGD} \\
    \cmidrule(lr){2-4} \cmidrule(lr){5-7}
    & BB-Generator & T5-Generator & Classifier & BB-Generator & T5-Generator & Classifier \\
    \midrule
         Batch size & \multicolumn{6}{c}{32} \\
         Learning rate & \multicolumn{6}{c}{5e-5} \\
         Learning rate decay & \multicolumn{6}{c}{linear} \\
         Warmup steps & \multicolumn{6}{c}{100} \\
         \midrule
         Epochs & 1 & 5 & 10 & 1 & 5 & 5 \\
         \midrule
         \textit{LoRA settings} & \\
         r & \multicolumn{2}{c}{8} & -- & \multicolumn{2}{c}{8} & -- \\
         alpha & \multicolumn{2}{c}{32} & -- & \multicolumn{2}{c}{32} & -- \\
         dropout & \multicolumn{2}{c}{0.1} & -- & \multicolumn{2}{c}{0.1} & -- \\
         \bottomrule
    \end{tabular}
    \caption{Training hyperparameters for the BlenderBot \citep{Roller2020RecipesFB} and T5 \citep{Raffel2019ExploringTL} generators and DistilBERT \citep{Sanh2019DistilBERTAD} classifiers.}
    \label{tab:hyperparameters}
\end{table*}

\section{Human Evaluation}
\label{sec:amt}
For human evaluation, workers were provided with the the following short instructions: \textit{You will be shown a target reply and two candidate replies. Select which candidate reply is most similar to the target reply.} They were also provided with the following long instructions: \textit{You will be shown a target reply and two candidate replies. Select which candidate reply is most similar to the target reply. Similar means having the same semantic meaning.}

\section{Disclosure}
GitHub Copilot was used to a limited extent for boilerplate code autocompletion. All models and the dataset used in this paper are freely available for use in research.

\end{document}